\documentclass[runningheads]{llncs}
\usepackage[T1]{fontenc}
\usepackage{booktabs}
\usepackage{amssymb}
\usepackage{xcolor}
\usepackage{hyperref}
\usepackage{amsmath}

\let\std\relax
\usepackage{multirow}
\usepackage{graphicx,verbatim}
\usepackage{subcaption}
\usepackage{placeins}
\usepackage{caption}
\usepackage{xcolor, soul}

\usepackage{tikz}
\usetikzlibrary{arrows.meta,positioning,fit}
\newcommand{\std}[1]{\text{\scriptsize$\pm$#1}}
\begin{document}
\title{Response-Aware Multimodal Learning for Post-Treatment Visual Acuity Forecasting}
\titlerunning{Response-Aware Multimodal Learning for Post-Treatment VA Forecasting}
%

\author{
  Phuoc-Nguyen Bui \inst{1} 
  Van-Vi Vo \inst{1} \and
  Duc-Tai Le \inst{2} \and
  Junghyun Bum \inst{3} \and
  Van-Nguyen Pham \inst{1} \and
  Kiyoung Kim \inst{4} \and
  Seung-Young Yu \inst{4} \and
  Hyunseung Choo \inst{5}\thanks{Corresponding author}
}
\authorrunning{Bui et al.}
\institute{
  Research Convergence Institute, Sungkyunkwan University, Korea\\
  \and
  Dept. of AI Systems Engineering, Sungkyunkwan University, Korea\\
  \and
  College of Computing and Informatics, Sungkyunkwan University, Korea\\
  \and
  Dept. of Ophthalmology, Kyung Hee University Medical Center, Korea\\
  \and
  Dept. of Electrical and Computer Engineering, Sungkyunkwan University, Korea
}

\maketitle           
\begin{abstract}
Long-term visual acuity (VA) outcomes after anti-VEGF therapy are central to patient counseling, expectation setting, and follow-up planning in diabetic macular edema (DME). However, in clinical practice, physicians must often estimate long-term visual trajectories based only on early post-treatment findings, making reliable prognostication difficult. Although prior OCT-based learning approaches have largely focused on short-term response or single-endpoint prediction, modeling VA trajectories across multiple future time points from early longitudinal observations remains insufficiently explored.
In this study, we assembled a real-world cohort of 188 anti-VEGF--treated DME patients with paired baseline and month-1 OCT scans, along with tabular OCT-derived biomarkers and non-imaging clinical variables. Using only these early data, we formulate a multi-horizon VA forecasting problem aimed at predicting visual outcomes at 3, 6, 12, 18, and 24 months, reflecting clinically meaningful follow-up intervals.
We propose \textbf{ReVA}, a response-aware multimodal framework that integrates structural features from baseline and month-1 OCT with the tabular variables to capture baseline disease status and early treatment response. ReVA uses spatial attention to preserve localized prognostic imaging features and a dependency-aware tabular encoder to model interactions among clinical variables. These multimodal representations are fused to predict patient-specific long-term visual acuity trajectories.
The proposed framework achieves MAE $=0.1246$, RMSE $=0.1621$, and $R^2=0.6064$ for 24-month VA prediction, with consistent performance across all forecast horizons. Our findings show that incorporating early treatment-response signals enables clinically meaningful long-term visual acuity forecasting, supporting data-driven decision support for routine anti-VEGF management. Code and pretrained models will be released on \url{GitHub}.


\end{abstract}


\section{Introduction}

Visual acuity (VA) is the principal functional endpoint in macular disease and directly informs anti-VEGF treatment decisions, follow-up scheduling, and patient counseling. Optical coherence tomography (OCT) is routinely acquired at high resolution, enabling learning-based models that predict VA or treatment outcomes from image data, often augmented with tabular variables. Yet robust multi-horizon VA forecasting remains difficult in real-world cohorts due to heterogeneous baseline status, variable treatment response trajectories, structure--function discordance, and pervasive missing follow-up and label noise.

Existing approaches broadly span (i) classical machine learning on structured tabular variables for short- to mid-term prognosis~\cite{rohm2018predicting,chandra2024predicting}, and (ii) deep learning that extracts prognostic representations from OCT, via either engineered imaging biomarkers or end-to-end modeling for cross-sectional and future outcomes~\cite{romo2020end,fu2021predicting}. Multimodal variants further combine OCT with tabular variables to improve post-treatment VA and anatomical prediction~\cite{leng2024development}, and attention-based or generative methods have been explored for response characterization~\cite{moon2023prediction}. Despite steady progress, two limitations constrain clinical deployment: most models target a single endpoint or short-term response rather than forecasting multiple future horizons, and many OCT encoders compress scans into global thickness/fluid-driven descriptors, underutilizing localized microstructural cues (e.g., outer retinal and photoreceptor integrity) that more directly bound achievable vision.

We study long-term VA prognosis in a clinically practical two-visit setting that explicitly leverages early treatment response. We curate a real-world cohort of 188 anti-VEGF--treated DME patients with paired baseline and month-1 OCT scans, together with tabular OCT-derived biomarkers and non-imaging clinical variables, and predict VA at 3, 6, 12, 18, and 24 months using only these early data. Our premise is that the change from baseline to month-1 encodes patient-specific treatment sensitivity and latent disease dynamics, providing a strong signal for long-range forecasting when represented with appropriate inductive bias. To this end, we propose a response-aware multimodal architecture that adapts a foundation-pretrained OCT transformer via LoRA, aggregates patch tokens with multi-head spatial attention to retain localized prognostic patterns, encodes tabular variables with a dependency-aware tabular module, and performs cross-attention fusion to obtain visit-specific multimodal representations. A lightweight temporal aggregation module then composes baseline state and early response into a compact embedding for multi-output VA prediction using temporal convolutional networks (TCN). In summary, our main contributions are three-fold:\sloppy
\begin{itemize}
    \item We curate a real-world two-visit DME cohort and formulate an early-response, multi-horizon forecasting task to predict VA at 3--24 months from baseline and month-1 data.
    \item We develop \textbf{ReVA}, a response-aware multimodal framework that fuses paired OCT and tabular variables and explicitly models early post-treatment change for long-range VA forecasting.
    \item We achieve strong multi-horizon forecasting performance, with best month-24 results of MAE $=0.1246$, RMSE $=0.1621$, and $R^2=0.6064$, and validate key components via ablation studies.
\end{itemize}

\section{Method}
\label{sec:method}

\begin{figure*}[!t]
    \centering
    \includegraphics[width=\linewidth]{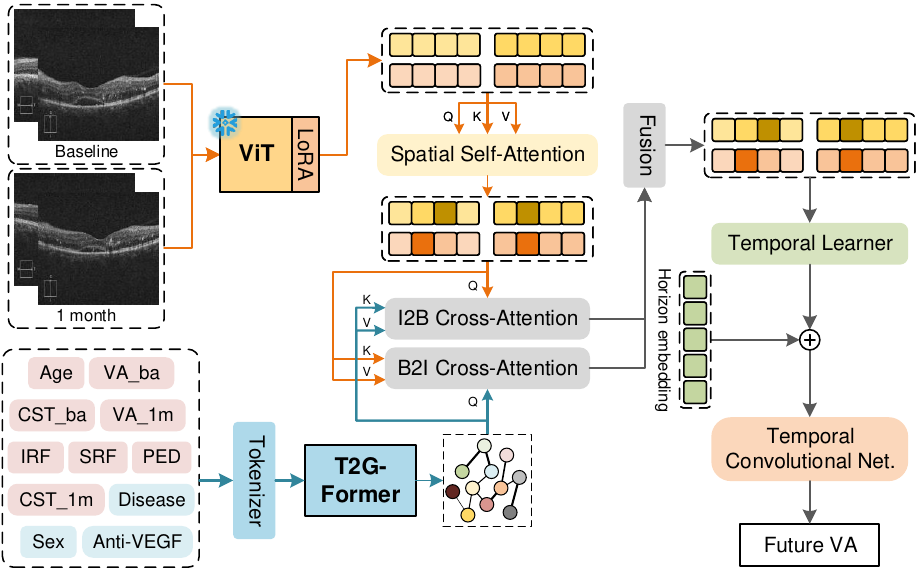}
    \caption{Overview of the proposed response-aware multimodal framework for multi-horizon post-treatment VA forecasting from two-visit OCT (baseline and month-1) scans and tabular variables. The model extracts OCT structural features with a LoRA-adapted foundation transformer, encodes tabular variables with a dependency-aware encoder, fuses the two branches via cross-attention, and aggregates the two visits to predict VA at multiple future horizons.}
    \label{method}
\end{figure*}

\subsection{Problem Formulation}
\label{subsec:problem}

For each patient, we observe OCT scans from two visits (baseline and month-1) with two orthogonal scan views (horizontal/vertical), together with tabular OCT-derived biomarkers and non-imaging clinical variables. The OCT input is
\begin{equation}
\mathbf{I}=\{\mathbf{I}^{(0)}_{H},\mathbf{I}^{(0)}_{V},\mathbf{I}^{(1)}_{H},\mathbf{I}^{(1)}_{V}\},
\end{equation}
where superscripts $(0)$ and $(1)$ denote baseline and month-1. Let $\mathbf{c}$ denote the tabular variables, consisting of numerical features (${\text{Age}, \text{ CST}_{\text{base}}, \text{ CST}_{\text{1m}}, \text{ VA}_{\text{base}}, \text{ VA}_{\text{1m}}}$) and categorical features (${\text{Sex}, \text{ IRF}, \text{ SRF}, \text{ PED}, \text{ Disease}, \text{ Anti-VEGF type}}$). Our goal is multi-horizon regression of future VA:
\begin{equation}
f_{\theta}:\,(\mathbf{I},\mathbf{c}) \mapsto \hat{\mathbf{y}}\in\mathbb{R}^{T},
\end{equation}
where $\hat{y}_t$ is the predicted VA at horizon $t\in\{1,\dots,T\}$ (we use $T{=}5$ in our implementation). When follow-up labels are missing, we use an observation mask $\mathbf{m}\in\{0,1\}^{T}$ to compute loss only over available horizons.

\subsection{Foundation OCT Encoding and Tabular Tokenization}
\label{subsec:foundation_graph}

Each patient provides two visits (baseline $t{=}0$ and month-1 $t{=}1$) and two orthogonal B-scan views (horizontal/vertical). Each scan is encoded by a RETFound-initialized ViT-Large, producing a CLS token and $N$ patch tokens:
\begin{equation}
\mathbf{Z}=[\mathbf{z}_{\mathrm{CLS}},\mathbf{z}_1,\dots,\mathbf{z}_N]\in\mathbb{R}^{(N+1)\times D}.
\end{equation}

To enable parameter-efficient fine-tuning, we inject LoRA~\cite{hu2022lora} adapters into attention projections (e.g., \texttt{qkv}, \texttt{proj}) by reparameterizing
\begin{equation}
\mathbf{W}'=\mathbf{W}+\Delta\mathbf{W},\qquad \Delta\mathbf{W}=\frac{\alpha}{r}\mathbf{B}\mathbf{A},
\end{equation}
with rank $r$ and scaling $\alpha$. When used, the backbone can be frozen so optimization is restricted to adapter parameters.
Given patch tokens $X\in\mathbb{R}^{B\times N\times D}$, we apply multi-head self-attention pooling to obtain a single view descriptor while retaining spatial attributions:
\begin{equation}
A^{(h)}=\mathrm{softmax}\!\left(\frac{Q^{(h)}{K^{(h)}}^{\top}}{\sqrt{d}}\right),\qquad
\tilde{X}=\mathrm{LN}\!\big(X+\mathrm{Proj}(\mathrm{Concat}_h(A^{(h)}V^{(h)}))\big),
\end{equation}
where $d=D/H$. The view embedding is computed by token pooling,
\begin{equation}
\mathbf{v}=\frac{1}{N}\sum_{n=1}^{N}\tilde{\mathbf{x}}_n\in\mathbb{R}^{D}.
\end{equation}

We represent tabular variables (numerical and categorical features) as a token sequence with a learnable readout token,
\begin{equation}
\mathbf{T}=[\mathbf{t}_{\mathrm{CLS}},\mathbf{t}_1,\dots,\mathbf{t}_L]\in\mathbb{R}^{(L+1)\times d}.
\end{equation}
Numerical features are embedded by feature-wise scaling of learnable token vectors, whereas categorical ones are embedded via lookup tables. We then encode the tabular tokens using T2G-FORMER~\cite{yan2023t2g} to capture higher-order feature interactions, and use the final readout token as the clinical embedding $\mathbf{g}\in\mathbb{R}^{d}$.

\subsection{Cross-Modal Fusion and Delta-Gated Temporal Learner}
\label{subsec:fusion_delta}

Given OCT view descriptors $\mathbf{v}_k$ and a clinical embedding $\mathbf{g}$, we obtain view-wise multimodal features via cross-attention fusion:
\begin{equation}
\mathbf{f}_k=\mathrm{Fuse}(\mathbf{v}_k,\mathbf{g}),\qquad k\in\{1,\dots,4\}.
\end{equation}

We then concatenate horizontal/vertical features within each visit,
\begin{equation}
\mathbf{b}=[\mathbf{f}^{(0)}_{H};\mathbf{f}^{(0)}_{V}],\qquad
\mathbf{m}=[\mathbf{f}^{(1)}_{H};\mathbf{f}^{(1)}_{V}],
\end{equation}
and summarize early response with a gated delta update:
\begin{equation}
\mathbf{s}=f(\mathbf{b}),\quad \Delta=g(\mathbf{m}-\mathbf{b}),\quad
\boldsymbol{\gamma}=\sigma\!\big(h([\mathbf{s};\Delta;|\Delta|;\|\Delta\|_2])\big),\quad
\mathbf{z}=\mathbf{s}+\boldsymbol{\gamma}\odot\Delta.
\end{equation}
Here $f,g,h$ are MLPs and $\boldsymbol{\gamma}$ adaptively suppresses noisy month-1 changes. We further apply \emph{delta dropout} by randomly zeroing $(\mathbf{m}-\mathbf{b})$ with probability $p$ during training.

\subsection{Time-Conditioned TCN for Multi-Horizon VA Regression}
\label{subsec:tcn_regression}

Given the temporally aggregated patient representation $\mathbf{z}$, we predict VA at $T$ future horizons by casting multi-horizon forecasting as sequence regression over \emph{ordered horizons}. This enables the predictor to share information across horizons and to model smooth, structured progression patterns without enforcing a rigid parametric trajectory. We first compute a shared latent descriptor
\begin{equation}
\mathbf{h}=\psi(\mathbf{z})\in\mathbb{R}^{H},
\end{equation}
where $\psi(\cdot)$ is an MLP. To specialize the prediction for each horizon $t\in\{0,\dots,T-1\}$, we add a learnable horizon embedding $\mathbf{e}_t\in\mathbb{R}^{H}$ and form a horizon-conditioned token as follows:
\begin{equation}
\mathbf{s}_t=\mathbf{h}+\mathbf{e}_t,\qquad \mathbf{S}=[\mathbf{s}_0,\dots,\mathbf{s}_{T-1}]\in\mathbb{R}^{T\times H}.
\end{equation}
This construction yields a compact horizon sequence $\mathbf{S}$ whose tokens share patient context ($\mathbf{h}$) while remaining horizon-aware via $\mathbf{e}_t$.

We then model inter-horizon dependencies with a dilated causal TCN, which captures both short-range coupling (nearby horizons) and long-range effects (distant horizons) using exponentially increasing receptive fields:
\begin{equation}
\tilde{\mathbf{S}}=\mathrm{TCN}(\mathbf{S})\in\mathbb{R}^{T\times H}.
\end{equation}
The TCN is implemented as a stack of residual temporal blocks with causal, dilated $1$D convolutions and dropout, preserving the horizon order while preventing information leakage from later to earlier horizons.
A linear readout produces per-horizon predictions
\begin{equation}
\hat{\mathbf{y}}=\rho(\tilde{\mathbf{S}})\in\mathbb{R}^{T},
\end{equation}
where $\rho(\cdot)$ is a point-wise linear layer (equivalently a $1\times1$ temporal convolution). With ground-truth $\mathbf{y}\in\mathbb{R}^{T}$ and observation mask $\mathbf{m}\in\{0,1\}^{T}$, we optimize a masked regression objective
\begin{equation}
\mathcal{L}_{\mathrm{reg}}=\frac{\sum_{t=0}^{T-1} m_t\,\ell(\hat{y}_t,y_t)}{\sum_{t=0}^{T-1} m_t+\epsilon},
\end{equation}
where $\ell$ is $\ell_1$ (MAE loss) and $\epsilon$ ensures numerical stability.

\section{Experiments}
\subsection{Dataset and Evaluation Metrics}

We curated a retrospective real-world cohort of $N{=}188$ anti-VEGF--treated DME patients with paired OCT acquired at two visits (baseline and month-1), along with tabular OCT-derived biomarkers~\cite{bui2025multi} and non-imaging clinical variables. Each visit includes two orthogonal B-scan views (horizontal and vertical), yielding four OCT inputs per patient. All splits were performed at the \textbf{patient level} to prevent data leakage~\cite{bui2023multi}. OCT slices are resized to $224{\times}224$ and center-cropped to emphasize the foveal region. Numerical tabular variables are z-score normalized using training-fold statistics, while categorical ones are index-encoded. The primary prediction target is best-corrected VA at 24 months post-treatment 
initiation (logMAR). Intermediate horizons (3, 6, 12, and 18 months) were optionally used as auxiliary supervision during training. Performance was evaluated using mean absolute error (MAE), root mean squared error (RMSE), and coefficient of determination ($R^2$). All results are reported under 5-fold cross-validation as mean${\pm}$std across folds.


\subsection{Implementation Details}

Unless otherwise specified, we use RETFound ViT-Large as the OCT backbone (output dimension $D{=}768$) with frozen weights during training. The clinical encoder is a T2G-Former with token dimension $d{=}128$, three graph-attention layers, and four attention heads. Spatial attention uses four heads, and multimodal fusion uses a hidden dimension of 512. A GRU-based temporal aggregator models two-visit dynamics, followed by a TCN head with a time-embedding dimension of 16. Dropout is set to $p{=}0.1$. Models were optimized using Adam with early stopping based on validation MAE. All experiments were implemented in PyTorch and trained on NVIDIA A6000 GPU.

\subsection{Comparison with State-of-the-Art}

\begin{figure}[!h]
\centering
\includegraphics[width=\linewidth]{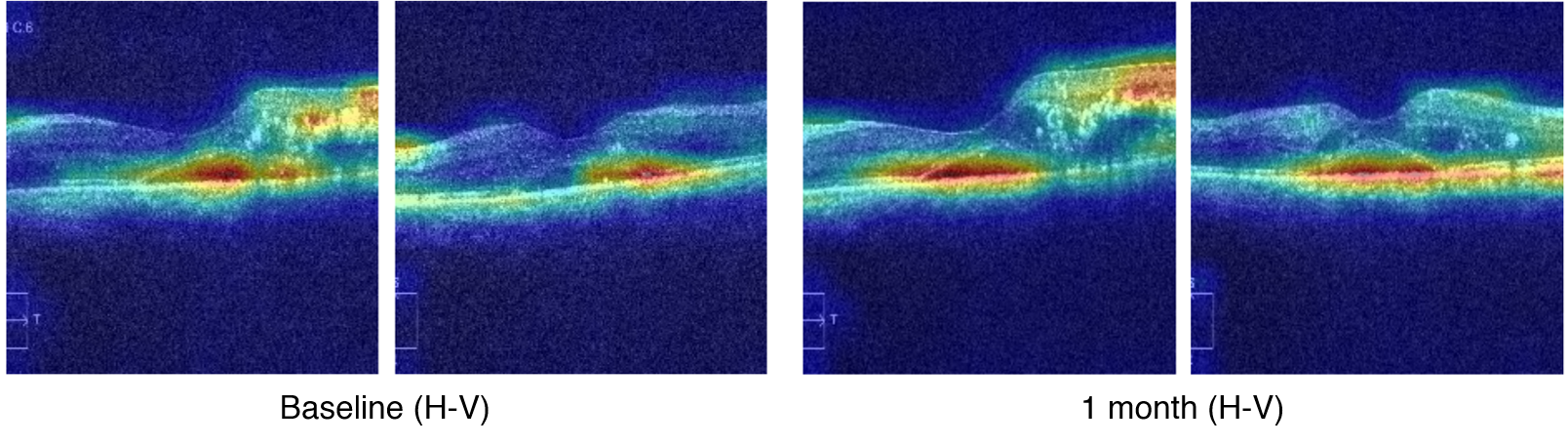}
\caption{Spatial attention heatmaps on representative OCT scans. Warmer colors indicate higher attribution to the predicted VA.}
\label{fig:heatmap}
\end{figure}

We compared our approach with representative CNN- and Transformer-based OCT backbones, as well as previously reported VA prognostic models. Quantitative results are summarized in Table~\ref{tab:sota}. Our model achieved the lowest MAE (0.1246), lowest RMSE (0.1621), and highest $R^2$ (0.6064). Compared with the strongest prior method, Rohm et al.~\cite{rohm2018predicting}, our approach reduced MAE by 10.17\% and improved $R^2$ by 9.80\% relative, with statistically significant differences ($p < 0.05$). From a clinical perspective, an MAE of approximately 0.12 logMAR corresponds to roughly one line on a standard VA chart. While a one-line difference may not independently alter treatment decisions, many clinical trials define meaningful visual change as a two- to three-line difference; thus, the primary value of our approach lies in improving long-term trajectory estimation and setting more informed expectations for patients. Beyond accuracy, spatial attention maps, shown in Fig.~\ref{fig:heatmap}, consistently highlight pathologically relevant regions, including subfoveal fluid and photoreceptor disruption. This agreement between model attribution and clinically recognized features improves interpretability and supports the biological plausibility of the predictions.

\begin{table}[!h]
\centering
\caption{Comparison of 24-month VA regression (logMAR). Lower MAE/RMSE and higher $R^2$ indicate better performance (\underline{underline}: second best, \textbf{bold}: best).}
\renewcommand{\arraystretch}{1} 
\setlength{\tabcolsep}{4pt} 
\resizebox{\textwidth}{!}{%
\begin{tabular}{l|ccc}
\toprule
Method & MAE $\downarrow$ & RMSE $\downarrow$ & $\text{R}^2$ $\uparrow$ \\
\midrule
Romo-Bucheli et al.~\cite{romo2020end} & 0.2358\std{0.0199} & 0.2806\std{0.0252} & 0.1552\std{0.0802} \\
Ko et al.~\cite{ko2022deep} & 0.1427\std{0.0147} & 0.1895\std{0.0143} & 0.4591\std{0.1254} \\
Leng et al.~\cite{leng2024development} & 0.1536\std{0.0201} & 0.1911\std{0.0227} & 0.4633\std{0.0785} \\
Han et al.~\cite{han2024anti} & 0.1657\std{0.0240} & 0.2091\std{0.0353} & 0.3541\std{0.1545} \\
Holste et al.~\cite{holste2024harnessing} & 0.1705\std{0.0167} & 0.2117\std{0.0236} & 0.3388\std{0.0986} \\
Kim et al.~\cite{kim2024prediction} & 0.1588\std{0.0226} & 0.1972\std{0.0249} & 0.4275\std{0.0965} \\
Anderson et al.~\cite{anderson2025enhancing} & 0.1486\std{0.0166} & 0.1874\std{0.0154} & 0.4809\std{0.0666} \\
Rohm et al.~\cite{rohm2018predicting} & \underline{0.1387\std{0.0056}} & 0.1741\std{0.0083} & 0.5470\std{0.0762} \\
Mondal et al.~\cite{mondal2025application} & 0.1396\std{0.0064} & \underline{0.1729\std{0.0114}} & \underline{0.5496\std{0.1002}} \\
\midrule
\textbf{ReVA} & \textbf{0.1246\std{0.0066}} & \textbf{0.1621\std{0.0056}} & \textbf{0.6064\std{0.0727}} \\
\bottomrule
\end{tabular}
}
\label{tab:sota}
\end{table}

\FloatBarrier

\begin{table*}[!h]
\centering
\caption{Ablation studies for 24-month VA regression (logMAR).}
\label{tab:ablation}
\begin{subtable}{\textwidth}
\centering
\caption{Impact of different input modalities on prediction performance.}
{\footnotesize
\setlength{\tabcolsep}{8pt} 
\begin{tabular}{lccc}
\toprule
Model & MAE $\downarrow$ & RMSE $\downarrow$ & $\text{R}^2$ $\uparrow$ \\
\midrule
OCT-only          & 0.1571\std{0.0163} & 0.1957\std{0.0179} & 0.4354\std{0.0675} \\
Tabular-only     & 0.1298\std{0.0073} & 0.1651\std{0.0589} & 0.5916\std{0.0734} \\
Multi-modal & \textbf{0.1246\std{0.0066}} & \textbf{0.1621\std{0.0056}} & \textbf{0.6064\std{0.0727}} \\
\bottomrule
\end{tabular}}
\end{subtable}
\hfill
\begin{subtable}{\textwidth}
\centering
\caption{Comparison of clinical tabular variables and OCT image encoders.}
{\footnotesize
\setlength{\tabcolsep}{4pt}
\begin{tabular}{lccc}
\toprule
Model & MAE $\downarrow$ & RMSE $\downarrow$ & $\text{R}^2$ $\uparrow$ \\
\midrule
MLP          & 0.1532\std{0.0196} & 0.1900\std{0.0208} & 0.4685\std{0.0725} \\
TabTransformer~\cite{huang2020tabtransformer} & 0.1329\std{0.00684} & 0.1693\std{0.0113} & 0.5745\std{0.0567} \\
T2G-FORMER~\cite{yan2023t2g} & \textbf{0.1246\std{0.0066}} & \textbf{0.1621\std{0.0056}} & \textbf{0.6064\std{0.0727}}  \\
\midrule
Swin-T~\cite{liu2021swin}       & 0.1453\std{0.0161} & 0.1828\std{0.0123} & 0.5038\std{0.0691} \\
DenseNet~\cite{huang2017densely}  & 0.1448\std{0.0124} & 0.1822\std{0.0128} & 0.5124\std{0.0141} \\
ConvNeXt~\cite{liu2022convnet} & 0.1432\std{0.0149} & 0.1785\std{0.0128} & 0.5268\std{0.0703} \\
RETFound~\cite{zhou2023foundation} & \textbf{0.1246\std{0.0066}} & \textbf{0.1621\std{0.0056}} & \textbf{0.6064\std{0.0727}} \\
\bottomrule
\end{tabular}}
\end{subtable}
\hfill
\begin{subtable}{\textwidth}
\centering
\caption{Effectiveness of two-visit temporal aggregation strategies.}
{\footnotesize
\setlength{\tabcolsep}{5pt}
\begin{tabular}{lccc}
\toprule
Model & MAE $\downarrow$ & RMSE $\downarrow$ & $\text{R}^2$ $\uparrow$ \\
\midrule
Naive concat.       & 0.1296\std{0.0081} & 0.1668\std{0.0079} & 0.5951\std{0.0758} \\
Transformer-based         & 0.1287\std{0.0083} & 0.1646\std{0.0076} & 0.5924\std{0.0933} \\
GRU-based & \textbf{0.1246\std{0.0066}} & \textbf{0.1621\std{0.0056}} & \textbf{0.6064\std{0.0727}} \\
\bottomrule
\end{tabular}}
\end{subtable}
\end{table*}

\subsection{Ablation Studies}
\label{subsec:ablation}

We conducted ablation analyses under identical protocols, focusing on 24-month VA prediction in Table~\ref{tab:ablation}.
First, multimodal fusion outperformed both OCT-only and tabular variables-only models. Clinically, this suggests that structural OCT captures localized microstructural correlates of vision, whereas non-imaging tabular variables provide contextual information that may explain structure--function mismatch when OCT findings are subtle.
Second, modeling dependencies among clinical variables (T2G-FORMER~\cite{yan2023t2g}) consistently outperformed simpler tabular encoders, supporting the notion that clinical features do not act independently but interact in meaningful ways.
Among OCT encoders, RETFound~\cite{zhou2023foundation} yielded the strongest performance, consistent with the idea that foundation pretraining on retinal data enhances robustness in relatively small real-world cohorts with long-term outcome noise.
Finally, modeling early change (baseline $\rightarrow$ month-1) as a gated update improved performance over simple concatenation or generic sequence models. In clinical terms, explicitly modeling early treatment response appears more informative than treating visits as independent observations, reinforcing the importance of the early response signal in long-term prognosis.

\section{Conclusion}

We present a clinically practical two-visit framework for long-term visual prognosis after anti-VEGF therapy in DME, forecasting VA from 3 to 24 months using only baseline and 1-month OCT scans combined with tabular variables from a real-world cohort of 188 patients. By explicitly incorporating early treatment response through multimodal modeling, our approach achieved strong predictive performance (MAE = 0.1246, RMSE = 0.1621, $R^2$ = 0.6064). Beyond numerical improvements, the model provides trajectory-level estimates that may help clinicians anticipate disease course, individualize follow-up intensity, and support patient counseling. While further validation in larger and external cohorts is required, these findings suggest that response-aware multimodal learning offers a clinically meaningful step toward scalable, data-driven prognostic support in routine anti-VEGF care.



%
%
%
\bibliographystyle{splncs04}
\bibliography{miccai}

\end{document}